# An Empirical Study on Google Research Football Multi-agent Scenarios

Yan Song[1†]    He Jiang[2†]    Zheng Tian[3*]    Haifeng Zhang[1]    Yingping Zhang[4]

Jiangcheng Zhu[4]    Zonghong Dai[4]    Weinan Zhang[5*]    Jun Wang[6]

[1]Institute of Automation, CAS, Beijing 100190, China

[2]Digital Brain Lab, Shanghai 200001, China

[3]ShanghaiTech University, Shanghai 200001, China

[4]Huawei Cloud, Guian 550003, China

[5]Shanghai Jiao Tong University, Shanghai 200001, China

[6]University College London, London WC1E 6PT, UK

**Abstract:**    Few multi-agent reinforcement learning (MARL) research on Google Research Football (GRF) [1] focus on the 11v11 multi-agent full-game scenario and to the best of our knowledge, no open benchmark on this scenario has been released to the public. In this work, we fill the gap by providing a population-based MARL training pipeline and hyperparameter settings on multi-agent football scenario that outperforms the bot with difficulty 1.0 from scratch within 2 million steps. Our experiments serve as a reference for the expected performance of *Independent Proximal Policy Optimization* (IPPO) [2], a state-of-the-art multi-agent reinforcement learning algorithm where each agent tries to maximize its own policy independently across various training configurations. Meanwhile, we open-source our training framework Light-MALib which extends the *MALib* [3] codebase by distributed and asynchronized implementation with additional analytical tools for football games. Finally, we provide guidance for building strong football AI with population-based training [4] and release diverse pretrained policies for benchmarking. The goal is to provide the community with a head start for whoever experiment their works on GRF and a simple-to-use population-based training framework for further improving their agents through self-play. The implementation is available at https://github.com/Shanghai-Digital-Brain-Laboratory/DB-Football.

## 1    Introduction

Football is one of the most popular and long-lasting sports worldwide. In a football game, two teams of 11 players compete for victory by scoring more goals. This often requires high-level cooperation in both offense and defense. With its complex nature in agent control, competition and cooperation, football has also been regarded as one of the most challenging testbeds for multi-agent intelligence research. Our research is conducted on the football game engine proposed by Kurach et al. [1] named the Google Research Football (GRF), which is the only realistic simulator that is available for public research. It bears great similarity to well-known football games such as FIFA and Real Football, vividly modelling the dynamics of ball and player as well as their interaction such as passing, shooting, etc. Many previous studies have examined their algorithms on simple academy scenarios provided by the GRF engine, which only involves a small number of players and is restricted to a single offensive and defensive round. However, the more challenging multi-agent full-game scenario where two teams of 11 players compete against each other in 90 minutes[1], is far less explored due to the following major difficulties from the perspective of reinforcement learning:

   **1.** The sparsity of rewards;

   **2.** The long duration of game playing;

   **3.** The high stochasticity in the game state transition.

From the perspective of multi-agent system:

   **4.** Role or credit assignment to players;

   **5.** The exponential increase in policy search space resulted from multi-agent cooperation and competition;

To our knowledge, our work is the first to scale a multi-agent reinforcement learning algorithm to the full-game GRF scenario and successfully beat the strongest

---

| Paper | Research Topics | Testing Scenarios | Open-sourced |
|---|---|---|---|
| Li et al. [18] | diversity | *Academy 3_vs_1 with keeper, Academy counterattack hard, 3_vs_1 with keeper (full field)* | Yes |
| Niu et al. [19] | communication | *3_vs_2 scenario* | Yes |
| Roy et al. [20] | coordination | *3_vs_2 scenario* | No |
| Yang and Tang [21] | Reward shaping | *Run_to_score_with_keeper, Pass_and_shoot_with_keeper, Run_pass_and_shoot_with_keeper,* | No |
| Ruan et al. [22] | coordination | *3_vs_2, 3_vs_6, 5_vs_5 multi-agent scenarios* | Yes |
| Wang et al. [23] | Reward shaping | *5_vs_5 half-court offense* | No |
| Jiang et al. [24] | Exploration | *Academy 3_vs_1 with keeper, Academy 4_vs_2 with keeper, Academy counterattack hard* | No |
| Pu et al. [25] | Cooperation | *3_vs_2 scenarios* | No |
| Niu et al. [26] | Coordination | *11_vs_11 full-game* | No |
| Huang [27] | Imitation | *11_vs_11 full-game* | Only pre-trained models and testing scrips |
| **Ours** | **Benchmarking from scratch** | ***11_vs_11 full game*** | **Training framework, trained model and testing scripts** |

**Table 1** Academic research on GRF and ours

built-in AI with 1.0 difficulty by training from scratch. In particular, we contribute an empirical comparison of various training configurations against built-in AI in the full-game scenario and provide technical advice on how to build up even stronger AI through self-play by our population-based training framework.

Our goal is to provide the community with a head start for whoever attempts to experiment with their works on GRF and a simple-to-use population-based training framework to further improve their AI through self-play. We also believe our research will provide great insights into the study of scalability issues in multi-agent learning.

To summarize, we listed our main contribution as follows:

1. We demonstrate the scalability and effectiveness of Independent PPO by beating the built-in AI with 1.0 difficulty in the 11v11 full-game scenario and conduct thorough empirical studies on various training settings.

2. We further improve our agents by population-based self-play [4] in the GRF 11v11 scenario and obtain stronger policies of different playing styles that can serve as good initializations or baselines for future research.

3. We release our distributed population-based reinforcement learning framework Light-MALib, with diverse pretrained models and data analytical tools specifically targeting on GRF research.

## 2 Related Works on GRF

### 2.1 Reinforcement Learning

Techniques for Reinforcement Learning (RL) have matured rapidly over the last few years, yielding high-level performance in various tasks such as games [4,5,6,7], robotics [8], advertising [9], mobile networks [10], etc. Reinforcement learning methods search for optimal policies by maximizing the expected cumulated rewards collected by the agent while interacting with the external environment [11]. This can be done by learning to estimate the goodness of state-action pairs [11,12] and directly optimizing the policy function [13]. Regarding the number of controllable agents, reinforcement learning can be categorized into two classes, *Single-Agent RL* (SARL) with one controllable agent and *Multi-Agent RL* (MARL) [2,7,14] where multiple agents generate actions based on their observations either individually or cooperatively. Google Research Football includes both SARL and MARL scenarios and has been adopted as a testbed for AI competitions as well as academic research.

### 2.2 Online Competition on GRF

*Single-Agent RL (SARL) Google Research Football 2020 Kaggle competition* [15] was held by the famous Manchester City F.C., which attracted over 1,000 team participation in the single-agent 11v11 full-game scenario. In this task, one side of the team needs to control merely the designated player (one at each step) and all of its teammates are controlled by the built-in AI. Among all the participants, reinforcement learning methods have been widely used and the winner—*WeKick* [16] adopted the self-play reinforcement learning approach. This competition has largely boosted GRF research from an engineering aspect and some of the open-sourced SARL submissions have provided us with valuable suggestions on both feature engineering and action masking at an early stage. Recently, *MARL CoG 2022 GRF tracks* [17] were held at the *Conference of Games* on multi-agent RL settings (5v5 and 11v11 full-game scenarios) with an increase in the number of controllable players. Unfortunately, few teams share their implementations. This encourages us to open-source our works and provide guidance on multi-agent GRF tasks.

## 2.3 Academic Research on GRF

As a large-scale multi-agent RL testbed involving both competitive and cooperation tasks, GRF has been experimented with in many academic studies as listed in **Table 1**. Many of them test on relatively simpler academy scenarios. Among works in *11vs11* multi-agent scenarios, Niu et al. [26] measure the average reward of their method but do not solve the game; Huang [27] applies imitation learning on WeKick single-agent demonstrations and achieves a high win rate in the 11v11 multi-agent full-game scenario. However, their performances are largely based on prior experiences, such as the expert demonstration and the *action_set_v2* action space. The *action_set_v2* is an official 20-dimensional action space that includes an extra *built-in AI* action choice compared to the 19-dimensional default action set. This action choice is likely to select valuable action according to built-in logic and therefore often results in good-quality exploration. One of our experiments also showed that *action_set_v2* can indeed largely accelerate convergence at the early stage of training. Meanwhile, the baseline MAPPO algorithm they compare their methods to has terrible performance and this contradicts our experience.

In contrast, our work focuses on solving multi-agent football matches from scratch without any previous pretrained models and sticking to the default action set with each action choice corresponding to a unique move in a football game. To the best of our knowledge, we propose the first open benchmark for GRF 11v11 multi-agent full-game scenarios from scratch and the first public detailed implementation of agent training along with it. We also carry out population-based training which improves the general abilities of our policies and release our trained models.

In Section 3 we introduce our implemented algorithm. Section 4 contains details of our distributed framework. Section 5 contains the experimental results of MARL and Section 6 discusses population training which further improves the abilities of football AI.

# 3 Algorithm

## 3.1 Notation

Appendix A.1 lists the notations used in this paper.

## 3.2 Problem Formulation

The multi-agent football scenario can be formulated as a Decentralized Partially-observed Markov Decision Process (DEC-MDP) problem. Similar to the definition of a Markov Decision Process, a DEC-MDP is defined by a 7-tuple $(n, \mathcal{S}, \mathcal{A}, \mathcal{O}, R, P, \gamma)$, where:

- $n$ is the total number of agents;
- $\mathcal{S}$ is the state space and $s_t \in \mathcal{S}$ represents the state at timestep $t$;

- $\mathcal{O}$ is the joint observation space of all agents and $o_t^i \in \mathcal{O}^i$ represents the observation of agent $i$ at timestep t, where $\mathcal{O}^i$ is the observation space of agent $i$;
- $\mathcal{A}$ is the joint action space of all agents, $a_t^i \in \mathcal{A}_t^i$ is the action taken by agent $i$ at timestep $t$, where $\mathcal{A}^i$ is the action space of agent $i$;
- $P(s_t|s_{t-1}, A_{t-1})$ is the state transition probability that models the probability of state $s_t$ after agents take joint action $A_{t-1} = (a_{t-1}^1, ..., a_{t-1}^n)$ at the previous state $s_{t-1}$.
- $R$ is a set of reward functions and $r_t^i = R^i(s_t, A_t)$ is the reward received by agent $i$ after agents take joint action $A_t = (a_t^1, ..., a_t^n)$ at state $s_t$;
- $\gamma$ is the discount factor;

## 3.3 Backbone MARL Algorithm: Independent Proximal Policy Optimization (IPPO)

Proximal Policy Optimization (PPO) has been widely used in both single-agent tasks [13] and multi-agent tasks [2]. In this work, we use PPO with parameter-sharing, where we learn a policy $\pi_\theta$ and a value function $V_\phi$ (parametrized by $\theta$ and $\phi$ separately) shared by all agents. Specifically, $\pi_\theta(a_t^i|o_t^i)$ denotes the probability of action $a_t^i$ taken by agent $i$ after observing $o_t^i$ and $V_\phi(s_t)$ denotes the estimated return of state $s_t$. In the IPPO setting, each agent tries to maximize its own cumulative reward and the overall objective is:

$$\theta \leftarrow argmax_\theta \mathcal{J}(\theta) = \mathbb{E}_{a_t, s_t}\left[\sum_t \gamma^t R(s_t, a_t)\right] \quad (1)$$

According to the loss derivation of PPO [13], The policy loss of IPPO can be computed as:

$$\mathcal{L}(\theta) = \sum_{i=1}^n \mathbb{E}_{s \sim \rho_{\theta_{old}}, a \sim \pi_{\theta_{old}}}\left[\min\left(\frac{\pi_\theta(a^i|s)}{\pi_{\theta_{old}}(a^i|s)}\hat{A}_t, \\ clip\left(\frac{\pi_\theta(a^i|s)}{\pi_{\theta_{old}}(a^i|s)}, 1 - \epsilon, 1 + \epsilon\right)\hat{A}_t\right)\right] \quad (2)$$

where the clip operator ensures that the ratio $\frac{\pi_\theta(a^i|s)}{\pi_{\theta_{old}}(a^i|s)}$ is within the range of $[1 - \epsilon, 1 + \epsilon]$ to avoid large parameter updates. Compared to Equation 1, the IPPO policy loss $\mathcal{L}(\theta)$ add tricks of (1) importance sampling $\frac{\pi_\theta(a^i|s)}{\pi_{\theta_{old}}(a^i|s)}$ to correct off-policy searching [11]; (2) clipping operation $clip(\cdot)$ to enable parameter regularization and (3) *Generalized Advantage Estimation (GAE)* [28] $\hat{A}_t$ to achieve bias-variance trade-off. The advantage estimation $\hat{A}_t$ is defined as follows:

$$\hat{A}_t = \sum_{l=0}^{h} (\gamma\lambda)^l \delta_{t+l} \tag{3}$$
$$\delta_t = r_t(s_t, a_t) + \gamma V_\phi(s_{t+1}) - V_\phi(s_t)$$

where $\delta_t$ is the TD error at time step $t$ and we set $h$ such that $t + l$ reaches the end of the episode. Intuitively, both hyperparameters $\lambda$ and $\gamma$ determine bias-variance tradeoff and $\gamma$ also determines the length of horizon. Detailed explanations of GAE can be found in [28].

The value loss function of IPPO is formulated as a clipped version of the Mean Square Error (MSE) loss between the value function $V_\phi(\cdot)$ and the target state value $\hat{V}_t$:

$$\mathcal{L}^i(\phi) = \mathbb{E}_{s \sim \rho_{\theta_{old}}} \left\{ \min \left[ \left( V_\phi(s_t) - \hat{V}_t \right)^2, \right. \right.$$
$$\left. \left. \left( V_{\phi_{old}}(s_t) + clip\left( V_\phi(s_t) - V_{\phi_{old}}(s_t), -\epsilon, +\epsilon \right) - \hat{V}_t \right)^2 \right] \right\} \tag{4}$$

where $\hat{V}_t = \hat{A}_t + V_\phi(s_t)$ is the sum of the advantage estimate and the current state value, which we call the GAE return. By minimizing $\mathcal{L}^i(\phi)$ the value function $V_\phi$ approaches $\hat{V}_t$.

Additionally, an entropy regularization term $\mathcal{H}(\pi_\theta)$ of policy function $\pi_\theta$ is added to the policy loss (Equation 2) to boost exploration, a commonly-used technique in RL literature. The overall learning loss has become:

$$\mathcal{L}(\theta, \phi) = \sum_{i=1}^{n} \mathcal{L}^i(\theta) + \mathcal{L}^i(\phi) + \lambda_{entropy} \mathcal{H}(\pi_\theta) \tag{5}$$

where $\lambda_{entropy}$ determines the scale of the entropy term and a larger value indicates higher exploration. The pseudocode of PPO training is shown in **Algorithm 1**.

In practice, since players are homogeneous where the

---

**Algorithm 1: PPO training workflow**

**Input**: initial policy parameters $\theta_0$ of policy $\pi_\theta$ and value function parameters $\phi_0$ of value function $V_\phi$;

**for** $k = 0, 1, 2, ..., K$ **do**

    Pull batches of trajectories $\mathcal{B}_k = \{\tau_i\}_{i=1}^{b}$ collected by policy $\pi_k = \pi(\theta_k)$ from the remote data server.

    Compute GAE return $\{\hat{V}_t^i\}_{i=1}^{b}$ and advantage estimate $\{\hat{A}_i\}_{i=1}^{b}$ based on the current value function $V_{\phi_k}$.

    Update the policy by minimising the policy loss $\mathcal{L}(\theta_k)$ (**Equation 2**) via stochastic gradient ascent with Adam, receive new policy parameter $\theta_{k+1}$.

    Update the value function by minimizing the value loss $\mathcal{L}(\phi_k)$ (**Equation 4**) via stochastic gradient ascent with Adam, receive new value function parameter $\phi_{k+1}$.

**end**

**return** a trained policy parameter $\theta_K$ and trained value function parameter $\phi_K$

---

**Algorithm 1** PPO training workflow

---

observation and action space are identical, we set up a pair of policy function (actor) and value function (critic) and share parameters $\theta, \phi$ across all players. During training, we apply the Adam [29] optimizer with a learning rate of $5 \times 10^{-4}$ and a clipping parameter of $\epsilon = 0.2$. The detailed hyperparameter settings can be found in Appendix A.4.

## 3.4 Population-Based Training

We use IPPO as the backbone MARL algorithm to learn policies through reward maximization. To further improve the general abilities of the policy and make it less exploitable when facing strong opponents, we also implement population-based training which maintains a population of policies for one player/team in the game and updates the population based on genetic or evolutionary algorithms. In particular, we leverage the self-play mechanism [30] to explore the policy space where an agent learns stronger tactics by competing with itself. The improved policies are added to the population pool and can potentially be chosen as the opponent candidate in the future. As expected, competing with incrementally stronger opponents enhances the general abilities of the population [31].

We implement two population-based training methods, *Policy-Space Response Oracles* (PSRO) [30] which we utilize to rapidly build up a population of football AI at the early stage and *League Training* [7] at the late stage to improve general skills.

**Policy-Space Response Oracle**: In a two-team normal-form game [31] denoted as $(\Pi, U)$. $\Pi = (\Pi_1, \Pi_2)$ denotes the set of policy profiles containing strategies of both teams and $U: \Pi \rightarrow R$ is the utility function that evaluates the utility of a particular policy profiles. A payoff matrix $\mathcal{M}$ has every entry representing the utility of a particular policy profile $(\pi_i, \pi_j)$ and can be expanded once a new policy profile has been evaluated. The workflow of a PSRO trial is shown in **Fig.2** and is summarized as follows:

1. initialize the population with a policy set $\Pi^{init}$ and compute initial payoff table $\mathcal{M}$;

2. at each generation, select the mix-strategy opponent $\Pi^{mix}$ by computing the Nash equilibrium [33] on payoff table $\mathcal{M}$;

3. train a multi-agent policy controlling multiple players best responding to a mix-strategy opponent $\Pi^{mix}$ and obtain the trained policy set $\Pi^{br} = BR(\Pi^{mix})$;

4. add the new policy $\Pi^{br}$ for each team to the population and the expected utility obtained by the simulation to the payoff table $\mathcal{M}$;

5. move to step 2.

Since in GRF football games both teams are symmetric, two sides of the player can follow strategies from a shared population pool. The pseudocode of a PSRO trial is shown in **Algorithm 2** where we add the benchmark AI (built-in AI)

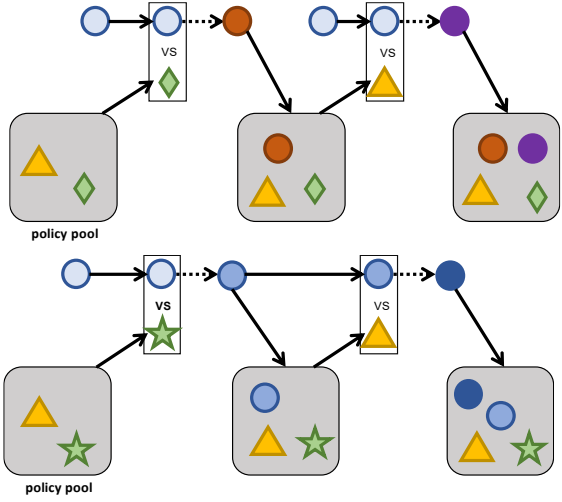

**Fig. 1** Top: PSRO with initializations and bottom: PSRO with inherits. Objects with various shapes indicate policies with different tactics.

to the population.

**Fig.1** shows schematic plots of two PSRO variants that differ mainly in best-response policy training (step 3). The first implementation of PSRO approximates a new best-response policy from scratch at each round. This often means that the policy needs to explore the game and gradually acquire basic skills such as dribbling the ball towards the opponent's goal repeatedly. A variant named *PSRO with inherits* inherit the best-response policy from the last round and continues learning against the new opponents in the current round. This largely saves computational time but suffers from the similar style-of-play among the trained policies.

In our implementation, we adopt the second version of PSRO and address the issue of limited diversity by setting up diverse reward shaping.

**League training**: League training is similar to PSRO but has a more complex framework that has been used to beat professional human players in StarCraft [7]. In league training, a main-agent is trained constantly to keep improving its ability (similar to PSRO with inherits) and the opponents are sampled following Prioritize Fictious Self-Play [7] (PFSP) which selects the opponents it cannot beat based on win rate. Thus, the goal of the main agent is to beat everyone in the population rather than a game theoretical subset of policies, enlarging the scope of opponent style-of-play and improving general abilities. In addition, an exploiter policy is reset at every generation and trained against the newest main agent only. Its job is to exploit the main-agent's strategic weakness and help to fix it by adding a copy of itself to the population. The exploiter can also potentially discover a new style-of-play as it does not inherit or only inherits from a general pretrained policy and has large space for evolving to counter the opponent. Furthermore, we drop the league exploiter to ease the computational burden.

The pseudocode of a league training trial is shown in

Algorithm 3. In practice, the training of exploiter-agents runs in parallel with the training of the main-agent.

In Section 5 we present empirical results of MARL training using the IPPO algorithm against built-in AI with difficulty 1.0 in GRF 11v11 full-game scenarios; In Section 6 we discuss how to use a population-based training framework to improve the general abilities of the policies

---

**Algorithm 2:** PSRO with benchmark AI included

**Input** a population $\Psi_\beta$ with size of $\beta$, a policy pool $\Pi$ with random policies $\pi_{\theta_1}$, Benchmark AI $\pi_b$, and other extra policies $\pi_e$;
randomly select two policies $\pi_l$ and $\pi_r$ as the initial policy of two teams (left team & right team)
**Initialize** $\alpha$ as the learning rate, $\gamma$ as the discounted factor, $\mathcal{G}$ as the max generation
**Initialize** $\mathcal{M}_{\beta \times \beta}$ as the payoff matrix for the meta game compute exp. utilities for each policy $\pi_i \in \mathcal{M}$, where $i \in \{1, \ldots, \beta\}$
**Result:** a set of best policies with different Elo rates
**for** $j$ *in* $[1, \ldots, \mathcal{G}]$ **do**
  **if** $j == 1$ **then**
    | set the opponent $\pi_o$ with Benchmark AI
  **end**
  **else if** $j > 1$ **then**
    | set the opponent $\pi_o$ with a probability $p$ sampled from Nash equilibrium policy in the payoff matrix $\mathcal{M}$
  **end**
  Train a multi-agent policy $\pi_j$ as the best response opposite to $\pi_o$ with IPPO
  Push it into the population $\Psi_\beta$, and update payoff matrix $\mathcal{M}$
**end**
**return** a set of best policies with different Elo rates

**Algorithm 2** PSRO workflow

---

**Algorithm 3:** League training

**Input** a population $\Psi_\beta$ with size of $\beta$, a main-agent policy $\pi_{m,0}$
**Initialize** $\alpha$ as the learning rate, $\gamma$ as the discounted factor, $\mathcal{G}$ as the max generation
**Initialize** $\mathcal{M}_{\beta \times \beta}$ as the payoff matrix for the meta game compute exp. utilities for each policy $\pi_i \in \mathcal{M}$, where $i \in \{1, \ldots, \beta\}$
**Result:** a main-agent policy
**for** $j$ *in* $[1, \ldots, \mathcal{G}]$ **do**
  Set the opponent $\pi_o$ with a probability $p$ following PFSP;
  Train the current main-agent policy $\pi_{m,j}$ as the best response opposite to $\pi_o$ with IPPO;
  Push the trained policy into the population $\Psi_\beta$ and update payoff matrix $\mathcal{M}$;
  **if** *main-agent finishes training* **then**
    Initialize an exploiter-agent policy $\pi_{e,j}$
    Set the opponent $\pi_{m,j}$;
    Train the exploiter-agent policy $\pi_{e,j}$ as the best response opposite to $\pi_{m,j}$ with IPPO;
    Push the trained policy to the population $\Psi_\beta$ and update payoff matrix $\mathcal{M}$;
  **end**
**end**
**return** current main-agent policy $\pi_{m,\mathcal{G}}$

**Algorithm 3** League training workflow

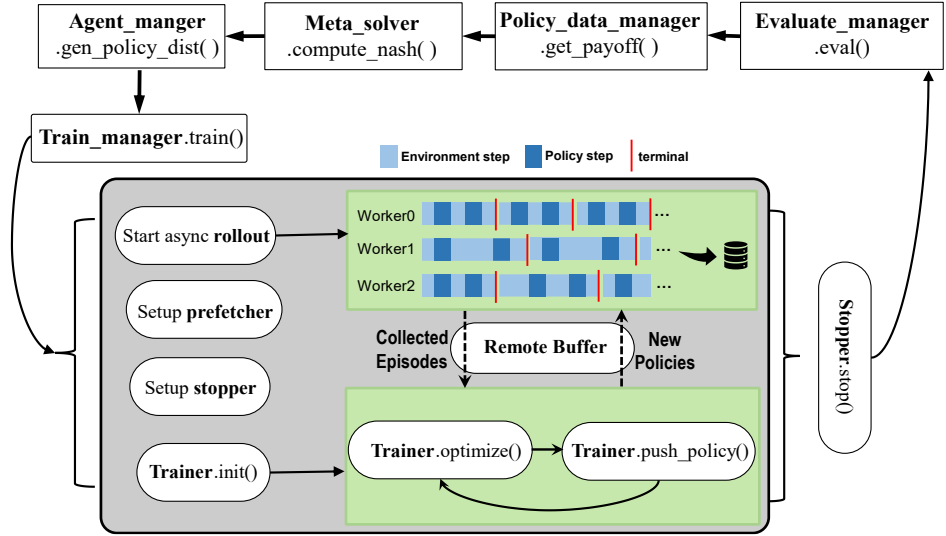

**Fig. 2** Workflow of a PSRO example in the Light MALib framework. The loop begins from an evaluation task (top right rectangle) and is followed by the construction of meta-strategies. The training manager activates trainers and generates rollout tasks with the given meta-strategies. The trainers iteratively carry out training and update policy to the remote buffer, while receiving the collected samples coming from the asynchronous rollout worker (similar to IMPALA [32]) through the buffer as well. The rollout worker renews its policy periodically. The process terminates according to some stopping criteria and proceeds to the next generation where the evaluation begins again.

and build strong football AI.

# 4 Distributed Framework: Light-MALib

To ensure consistency in the evaluation of MARL training, we open-source our Light-MALib (L-MALib) framework. As the name suggests, L-MALib is a simplified version of the MALib [3] codebase that uses *Ray* as a backend and offers population-based training pipelines (e.g., PSRO). L-MALib follows the design of *Actor-based* RL systems that execute algorithms through message passing between a set of actors deployed on worker nodes. Similar to MALib, L-MALib possesses a high-level agent/evaluator/learner abstraction but simplifies the workflow and optimizes efficiency to cater to large-scale training on GRF multi-agent scenarios. More comparisons between L-MALib and MALib can be found in Appendix A.2.

## 4.1 L-MALib Workflows

**Fig. 2** provides an overview of a PSRO trial using L-MALib in an evaluation-training/rollout-evaluation loop. The *Evaluation Manager* is responsible for simulation between policies in the pool. The *Policy Data Manager* and the *Meta-Solver* analyse the results and the *Agent Manager* generates the policy distribution for rollout tasks and the trainable policy id for training tasks. Inside the *Training Manager*, rollout and training tasks are executed asynchronously. The rollout workers receive a new policy from the remote buffer and return the collected episodes. The training workers constantly fetch samples from the buffer and renew the policy copy periodically.

On GRF specifically, L-MALib allows training MARL algorithms against built-in AI and a group of trained policies, as well as evolving a policy population by iterating between training and opponent selection. In addition to GRF-related content, L-MALib also provides a smaller-scale multi-agent RL environment benchmark for framework testing such as Leduc Poker [33], Kuhn Poker [31] and ConnectFour [34].

## 4.2 GRF Toolkits

To better evaluate algorithmic performance on football, we have also designed a football event detection data structure and developed a game debugger for detailed analysis per game. The new data structure as shown in **Fig.3** abstracts every change in ball ownership and records the involved players' status, such as the player who passes the ball and the player who receives it in a ball passing event. This enables us to accurately locate every game event such as a good assist or a bad pass and correspondingly assign individual credits. The game debugger loads from the official dump file of each game replay and provides elaborate 2D visualization of the game data as well as trajectory analysis for each individual player. This is beneficial for strategic analysis and helps identify weaknesses.

In brief, we present L-MALib, a population MARL distributed framework that inherits structural design from MALib but is specifically optimized for experiments on Google Research Football. We implement Independent Proximal Policy Optimization (IPPO) algorithms and test them on 11v11 multi-agent full-game scenarios.

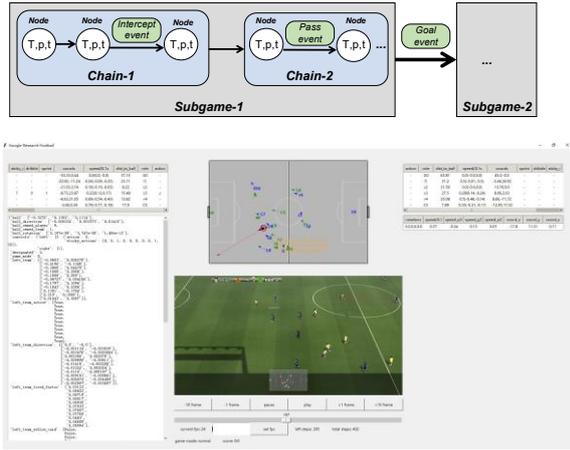

**Fig.3** Game graph and game debugger designed for GRF

## 4.3 Time Cost Estimate

Light-MALib speeds up the training phase by running environment rollouts and policy updates in a distributed and asynchronous manner. In addition, as we treat all players homogeneously by parameter-sharing, only a pair of actors and critics is maintained for each policy and hence the inference and update of networks can be performed in a single batch. During decentralized execution in a real game, each player has a copy of the trained policy and makes decisions based on his own observations. Specifically, at each decision step:

1. Each agent receives its own observation and encodes it into a high-dimensional feature vector;
2. Each agent feeds the feature vector into its actor network and decodes an action. The time cost is roughly the time of forward-passing of three-layer neural networks in our setting;
3. All actions are then collected and passed to the game engine. Finally, an environment step is executed and new observations are returned.

The time cost and relative proportion of each substep during rollout and training are shown in **Fig.4**. During policy execution, the GRF game engine takes 73% of the time and the batched policy inference only occupies 9.5% of the runtime. The maximum running speed is roughly 71 frames per second and on a 3000-step episode, the estimated time cost is approximately 1 minute. Thus, our AI can run in a real-time manner when deployed. During policy training, most of the time is spent on GPU loading and loss computation. Practically, the total runtime of our experiments is largely restricted by the policy execution and particularly the engine processing.

# 5. Empirical results of MARL Training

Through our experiments on GRF, we found several factors that are critical to performance such as the training resources, feature encoder, discount factor and reward shaping. In this section, we evaluate Independent PPO with the commonly-used setting on 11 vs 11 multi-agent scenarios. We apply only mild tuning and empirically investigate the following questions:

*1. How should feature encodings be chosen?*

*2. To what extent does discount factor γ affect performance under such a long-horizon task?*

*3. How does reward shaping affect performance in failure cases?*

*4. How many computational resources are required to achieve good performance within a reasonable amount of time?*

*5. How does the obtained policy behave?*

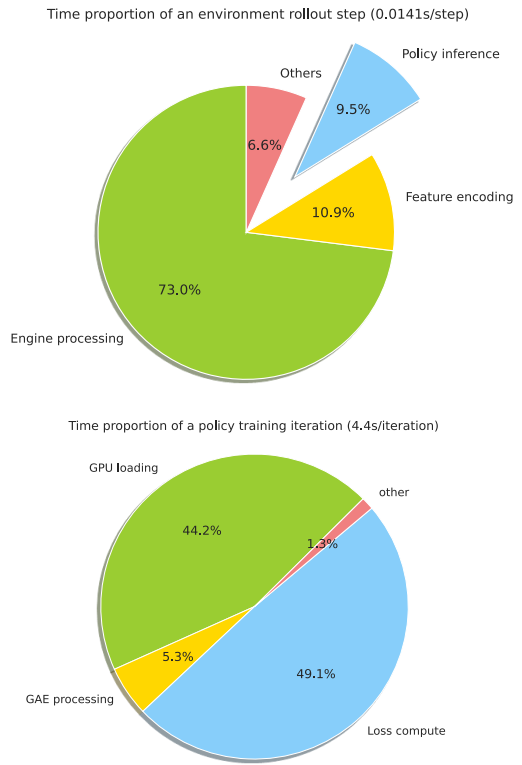

**Fig.4** Time proportion plot of environment rollouts (Top) and policy training iterations (Bottom). An environment rollout step takes approximately 0.014 seconds and most of the running time is used for engine processing. A training iteration takes approximately 5 seconds and most of the running time is used for GPU loading and loss computing.

The detailed experimental settings can be found in Appendix A.4, including the feature encoding, action masking, reward shaping and hyperparameter settings. All the empirical results are obtained from the following hardware settings:

*System*#1: A 128-core computing node with NVIDIA RTX3090 GPU (24G memory);

*System*#2: A 256-core computing node with NVIDIA A100 GPU (40G memory);

*System*#3: A 300-core computing node with 8x

NVIDIA 3080 GPUs (80 G memory in total)

## 5.1 Performance on Different Feature Encoding

We compare the performance of three different feature encodings: official *simple115_v2*, *encoder_basic* and *encoder_enhanced* as discussed in Appendix A.4. As the original *simple115_v2* encoder did not use action masking, we implement it with an action mask for a fair comparison. The results are shown in **Fig.5**. The vectorized *simple115_v2* performs terribly, indicating the need for extra information in the feature embedding. *Encoder_basic* performs the best by introducing states of closest teammates and opponents. The *encoder_enhanced* features have slower convergence. This is possibly due to the insignificance of extra information about the match states and off-side position when competing with built-in AI. However, when training in a population-based style, we resort for an enhanced encoder as we need more complicated features to counter stronger and more diverse opponents, such as using offside information to avoid off-side trapping.

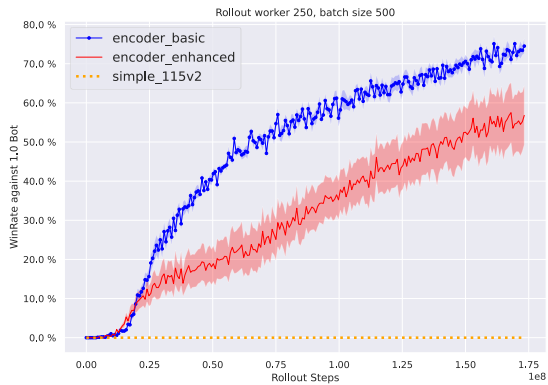

**Fig.5** Rollout win rate against the 1.0 hard bot with the official simple115_v2 (dashed line), encoder_basic (solid line with dot marker) and encoder_enhanced (solid line) feature encoders.

## 5.2 Performance on Different Discount Factors and Failure Cases

Our experiments also demonstrate that in 11v11 multi-agent scenarios, a discount factor $\gamma$ equal to 1 provides more stable training performance as illustrated in **Fig.6**, where $\gamma = 1$ indicates faster convergence and higher stability. The performance in each seed can be found in **Fig.19** where seed 4 is hard to learn with a lower discount factor. To explore why this happened we evaluate the learned state-value function $V_\phi$ on a testing trajectory and the results are drawn in **Fig.7**. We see that polices trained with $\gamma = 0.995, 0.99$ have a large variance $V_\phi$ across time steps and show greater aversion to goal loss as well as states that can potentially lead to it[2]. This might be due to the large TD error during bootstrapping as illustrated. The large TD error comes from the large scale in $V$ due to the non-zero discount factor and long-horizon which has also been noted in [28]. As a result, a large value in $\delta$ results in high variance in GAE returns across time (according to Equation 3), which further leads to high variance of the updated state-value estimate $V$. This renders the behavior of the actor pessimistic since the collected samples at the early training stage are often unrewarding over which $V$ shows great aversion. Additionally, in **Fig.20** we evaluate the same statistics on checkpoint models of both successful and failed policies throughout training. The policy that learns to defeat bots at $\gamma = 1$ manages to withhold its extreme aversion over unrewarding events and keep the value of $\delta$ small as the training proceeds, whereas the policy with $\gamma = 0.995$ founds it hard. Furthermore, **Fig.8** compares the values of $V$ and verifies that models trained with discount factors of 0.995 and 0.99 enlarge the scale of $V$ and make it pessimistic incrementally throughout training. Value clipping does not alleviate this issue. Therefore, we recommend using a high value of discount factor in the 11v11 full-game scenario.

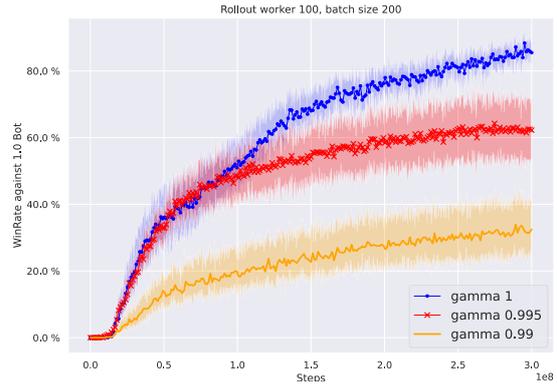

**Fig.6** Rollout win rate against 1.0 hard bot across different discount factors. The solid line with dot markers is at discount factor 1, the line with cross markers is at 0.995 and the solid line without markers is at 0.99. Results are averaged over four random seeds and the shaded area shows the standard error.

**Reward shaping in failure cases:** Previously we demonstrated that different discount factors result in different stabilities, and small values may lead to pessimistic behavior. Here, we fix the instability by reducing the negative rewards and investigate how it helps improve performance in failure cases. In **Fig.9**, we compare the performance of different reward shaping weighting at $\gamma = 0.995, seed = 4$, where zero-sum reward weighting (+1 for making a goal and -1 for losing a goal) does not work. After scaling down the weights of negative rewards assigned to goal losing events, the policy trained with $\gamma = 0.995$ manager to beat the bot. The fastest convergence occurs

---

[2] Reviewing the game replay, we found that the sharp drop in state-value estimates at the time-step where no goal/lose goal happens is a

consequence of opponent missing a shot.

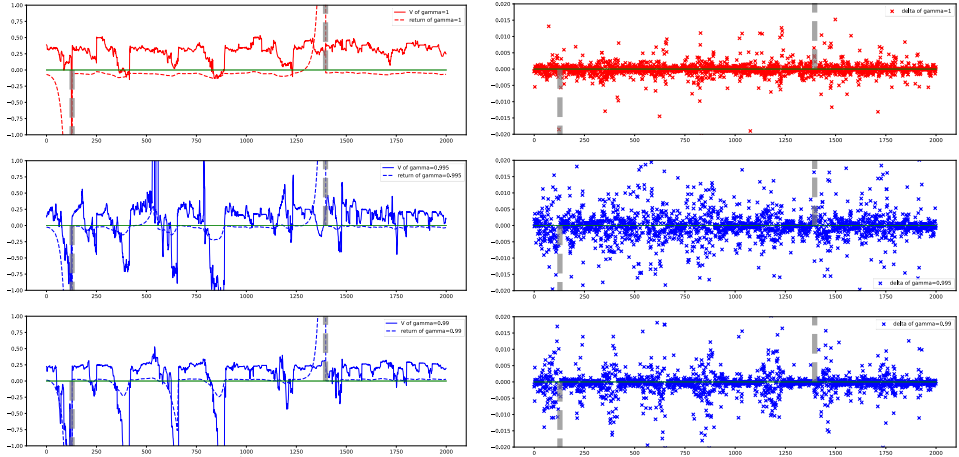

**Fig.7** State-value estimate (Left) and TD error (Right) of policy trained with discount factors of 1(Top), 0.995(Mid) and 0.99(Bottom) on a testing trajectory. The solid line indicates the state-value estimate $V$ and the dashed line indicates the GAE return across each time-step. The thick grey vertical bar represents the reward of goal (positive) and loss of goal (negative).

when the negative reward weighting is 0.2. This indicates that the negative rewards in GAE can potentially cause instability in long-horizon tasks when the discount factor $\gamma$ is low.

Meanwhile, this can also be due to the purely-offence style of plays from the trained policy where charging forwards is efficient to cause opponent offside and retake the ball ownership (as shown by the high win rate orange curve in **Fig. 9** where only positive rewards are considered).

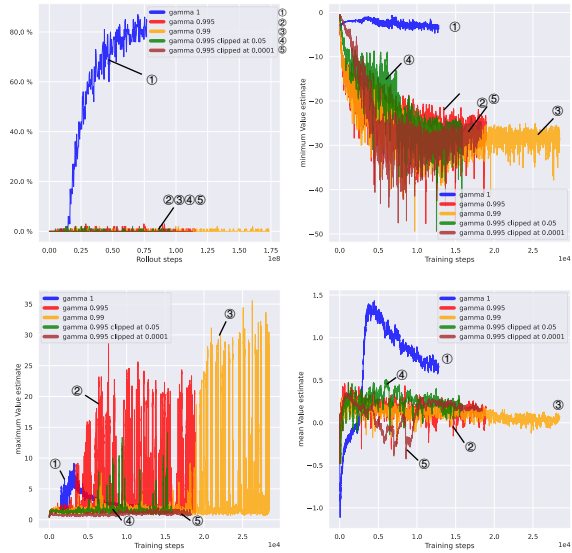

**Fig.8** Win rate and state-value estimate throughout training at different discount values and clipping thresholds (①: $\gamma = 1$; ②: $\gamma = 0.995$; ③: $\gamma = 0.99$; ④: $\gamma = 0.995$ with a 0.05 clipping threshold; ⑤: $\gamma = 0.995$ with a 0.0001 clipping threshold). All settings at $\gamma = 0.995$ cannot beat the bots and have large scales on $V$. Value clipping cannot alleviate such issues.

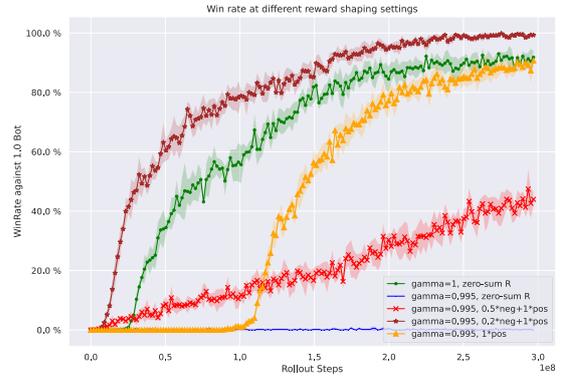

**Fig.9** Comparing the performance of different weights in reward shaping under random seed failure. The curves have been smoothed.

## 5.3 Performance on Different Computational Resources

Having determined a good feature encoder and the discount value. We evaluate the L-MALib performance of different resource configurations. The results can serve as a reference for researchers interested in 11v11 scenarios. Note that these experiments do not indicate parallelism performance but only provide a clue on how to configure training given available computing resources (on a single GPU machine or multi-GPU multi-node cluster). The factors we aim to test are as follows:

1. Number of rollout workers: number of parallel tasks that collect and return samples from environments. Each rollout work holds one environment task by default. The choices are [100(#1), 150(#2), 250(#3)][3].
2. Batch size: number of data sampled from the data server in each training iteration. The choices are [100(#1), 200(#1), 300(#2), 500(#3)].

We choose *encoder_basic* feature encoders and $\gamma = 1$. We run rollouts on the CPU and training on the GPU. The number of rollout workers determines the efficiency of sample generating on CPU machines, and the batch size determines the amount of sample used to compute the gradient per training iteration on GPU machines. We run experiments against the 1.0 hard bot with the same hyperparameter settings but different numbers of rollout workers and batch sizes on three systems. **Fig.10** illustrates the number of rollout iterations required to reach a good average win rate, where we see that GRF is a very resource-consuming testbed for IPPO, and within the capability of our machine, the performance becomes stably better given more resource allocation. In **Fig.11**, we also observe that our implemented algorithm is able to provide robust performance in a large scale setting.

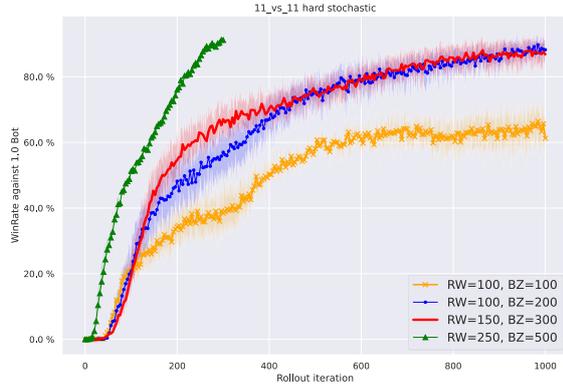

**Fig.10** Rollout win rate against a 1.0 hard bot across different rollout-batches configurations. The results are averaged over four random seeds and the shaded area represents the standard error. For each trial, the average time cost is approximately 70 seconds/iteration and the total time cost for 1k iterations is approximately 19 hours. At each rollout iteration, RW number of environment tasks are running in parallel and $RW \times 3000(Episode\ Length)$ steps of samples are collected. During training, $BZ \times 1000(Sample\ Length)$ steps of data are sent for training.

**Fig.19** explicitly shows the corresponding time performance at each seed, and **Table 2** lists the averaged time cost at a particular win rate. On System#1 and #2, our training framework costs approximately 15.7 hours to beat built-in AI with difficulty 1.0 from random initialization, and this time cost can be reduced to less than 5 hours, which is acceptable to use even on small-scaled clusters (such as System#1).

| | 20% | 40% | 60% | 80% |
|---|---|---|---|---|
| *RW=100,BZ=100(#1)* | 3.7 h | 5.1 h | 12.7 h | NA |
| *RW=100,BZ=200(#1)* | 4.2 h | 6.1 h | 9.7 h | 15.7h |
| *RW=150,BZ=300(#2)* | 3.1 h | 5 h | 8.8 h | 15.7h |
| *RW=250,BZ=500(#3)* | **0.6 h** | **1.6 h** | **3.3 h** | **4.9 h** |

**Table 2** Averaged time cost at particular win rate (20%,40%,60%,80%)

In brief, the carefully-designed feature encoders provide extra information and enable faster convergence. The high value of the discount factor ensures the stability of training and the larger computational resources offer higher efficiency and more robustness.

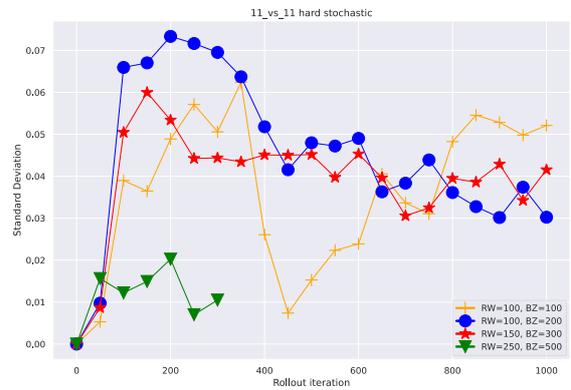

**Fig.11** Standard deviation of win rate against 1.0 hard bot across four random seeds at different resources configuration. Our implemented algorithms get robust performance at large scale.

## 5.4 Charging Forward is Efficient to Beat the 1.0 Hard Bot

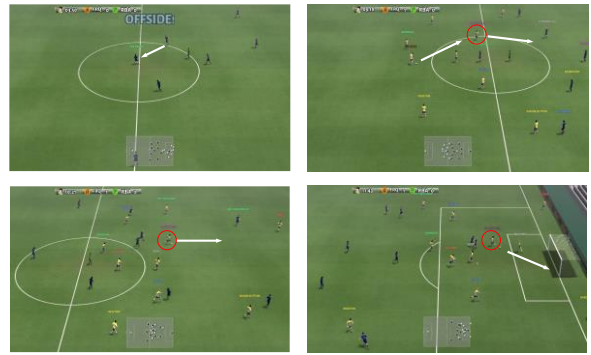

**Fig.12** The behavior of learned policy against hard bot. Hard bot can easily be offside when all of our players are in the right half (Top Left). We then start the free kick (Top Right) and pass to one of our teammates who charge forward recklessly (Bottom Left) and make a shot (Bottom Right).

Although we have achieved a fairly high win rate against the 1.0 hard bot, the style-of-play of all of our trained policies converges to a fixed offense and defense pattern demonstrated by snapshots of game replay in **Fig.12**. On defense all players controlled by the trained policy will move forwards towards the opponent's half to cause offside and dribble the ball all the way to the penalty area to make a shoot while attacking. This strategy works against bots surprisingly well since the built-in AI logic designed for hard bots sometimes does not recognize offside cases and has considerable defensive vulnerability. However, this charging strategy is also easily exploitable as it overfits to the glitch in bot.

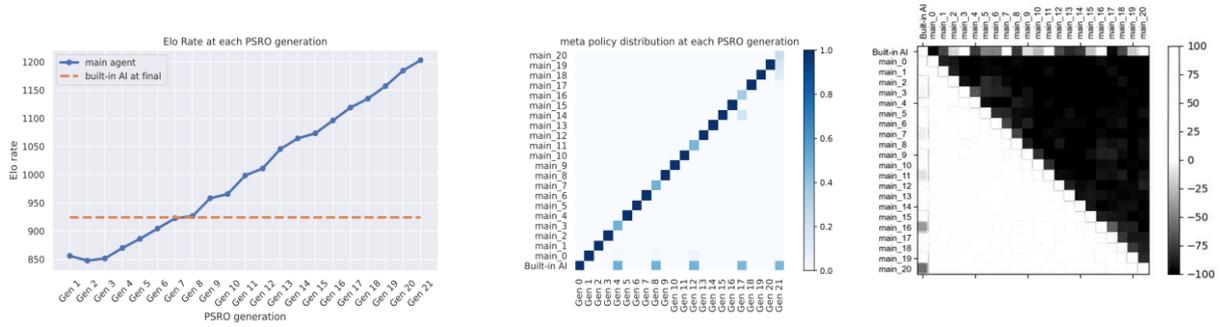

**Fig.13** PSRO on 11_vs_11 multi-agent scenarios. The left is the change in the Elo rate at each generation; the middle is the meta-policy distribution or the opponent strategy distribution at each generation. The right is the payoff table where each entry in the matrix represents the score of the policy on the vertical axis when competing with the policy on the horizontal axis. From the meta-distribution and the payoff table one can observe the non-transitivity in GRF training.

In other words, being able to beat the hard bot does not reveal strong general skills and we ought to use a more complicated framework, such as population-based training, to build general football AI.

# 6 Empirical Results of Population-Based Training

In Section 3.4 we have discussed our implemented population-based training methods. In practice, we build a general football AI through the following procedures:

**Stage 1**: Rapidly building up a pool of policies that are generally not strong but strategically diverse using diverse reward shaping;

**Stage 2**: Use the policy pool as diverse initializations and improve basics skills by carefully choosing opponents;

**Stage 3**: Gather representational policies and adopt a more complex evolutionary framework to improve general abilities.

## 6.1 Stage One: Increase the Reward Diversity

We first handcraft K skills in football games including offense and defense with different reward shaping functions. Then we train K populations using PSRO with inherits with K sets of reward functions $r_i(s_t, a_t)_{i=1}^{K}$. The best team policy with highest Elo rate in each trial would be selected as a representational policy at this particular reward shaping scheme. The detailed reward shaping and algorithm are shown in Appendix A.4.

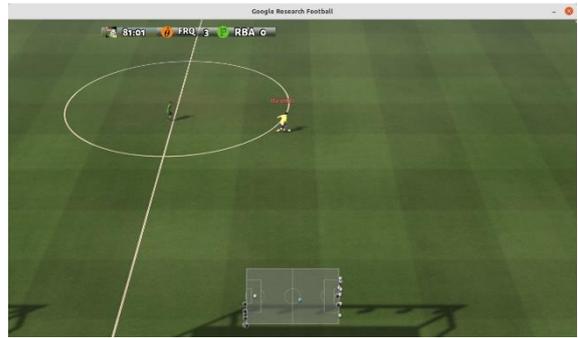

**Fig.14** A snapshot of the game replay between two policies during a PSRO run.

Note that these K number of policies might not be generally strong since tuning reward shaping towards a sophisticated football strategy is extremely hard and computationally expensive. Thus, at this stage we are not concerned about the strength of each policy as they are already the best along their evolutionary pathway, but the strategic diversity they express. For example, a trial using players' distance reward could learn a policy with a high-pressuring strategy while an active defensive reward shaping scheme could offer policies with stronger defense. Expectedly, we will have access to a set of representational policies with distinct strategical styles when stage one ends. However, in practice we encounter a failure case where we include built-in AI in the population of a PSRO trial. The trained policy inevitably sticks to a fixed strategy regardless of the reward-shaping scheme.

## 6.2 Fail Case: Experiments with Built-in AI Included

During our experiment, we found that adding built-in AI into the initial population can restrict the playing style of the resulting policy to a large extent. In **Fig.13**, we present the results of a PSRO run with a specific reward-shaping scheme. We adopt PSRO with inherits and for each generation, we stop training when the average win rate has exceeded 90% or the maximum number of rollout iterations has been reached. Built-in AI is included in the initial population. We apply the reward factors shown in **Table 3** with the following weighting:

$$5 \times (\textbf{team goal/lose goal}) + 0.2 \times (\textbf{individual goal and assist})$$
$$+ 0.05 \times (\textbf{individual lose ball})$$
$$+ 0.01 \times (\textbf{minimum distance})$$

Both the meta-strategy distribution and payoff table show the existence of non-transitivity [35] in this PSRO trial, which means that a policy that can beat another policy that beats the bot, loses to the bot. By reviewing the replay of the game, we found that the newest policy still has the same style-of-play as the policy in the first round which learns to beat the bot. A snapshot of the game replay is presented in **Fig.14**, where we see both teams standing at the opponent's penalty area and showing no defensive move. This is because the policy in the first round learns the charging style-of-play which we have mentioned in Section 5.4 and the following generation inherits this strategy and only updates slightly which is sufficient to beat the past, such as improving shooting accuracy or dribbling.

At the same time, a change in playing style can result in a large drop in the win rate against built-in AI since such a charging strategy is based on the strategic loophole that exists in built-in logic and any deviation from this strategy will be 'corrected' by gradient descent. In other words, the charging playing style is a local minima and is hard to escape from it. To verify this, we select policies *main_16* and *main_17* in **Fig.13**. *Main_16* gets beaten by the built-in AI (shown on payoff) but at generation 17, the converged policy *main_17* beat the built-in AI again since the mix-strategy of the opponent contains both bot and the past (shown on meta-strategy distribution). We look at the game replay from both policies and take a snapshot each at a defense round (shown in **Fig.15**). *Main_16* shows a defensive move while *main_17* gives up defending and takes advantage of the glitch in the bot (not recognizing the offside) to win the game. This happens because learning to defend with specially-designed rewards cannot easily stop the bot's mature attack and therefore the behavior rollbacks to the original strategy cause offside. This indicates that although defensive reward shaping encourages players to play defense, competing with built-in AI can waste such efforts.

In brief, the defensive glitch in built-in AI lures the trained policy to learn the strategy of charging while the strong general skills of built-in AI ensure that the trained policy does not deviate from this extreme playing style. As a result, we exclude built-in AI from the opponent pool and add policies with the highest ELO rate from other PSRO trials to the current initial population. This helps construct a population incrementally with policies that are not generally strong, but strategically diverse. In the next stage, we aim to improve the basic skills of the policies by carefully choosing the initialization and the opponent policies.

## 6.3 Stage Two: Initialization Diversity and Specific Opponent Selection

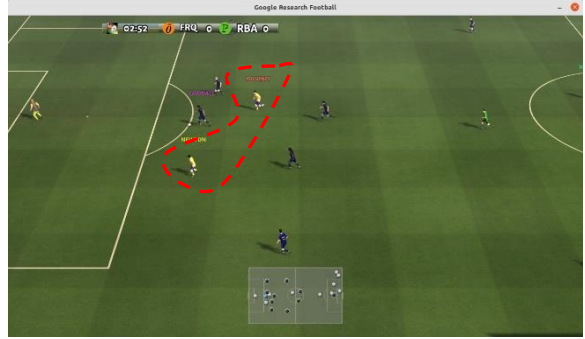

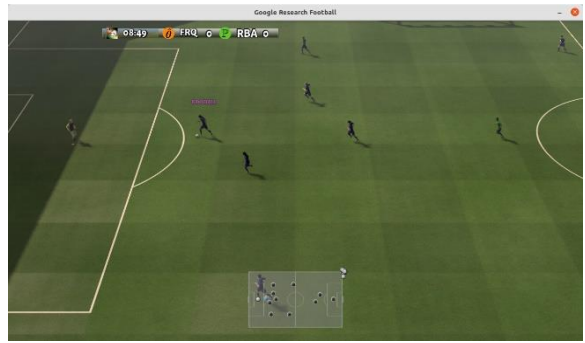

**Fig.15** Snapshots of a defensive round against bot from policy main_16 (Left, team yellow) and main_17 (right, team yellow). The main_16 has two defense players whereas main_17 gives up defending. *Main_17* can beat the bot while *main_16* cannot.

In all of our studies above, we consider how to train effective football policies and build policy populations from scratch. In addition, we provide further guidance on how to constantly improve the policy by using diverse policy initialization and selecting a specific opponent strategic style.

**Good initialization is important for training efficiency and policy diversity.** Training from scratch is computationally insufficient as the policy needs to acquire basic skills such as passing and shooting, which is difficult in such a sparse and long-horizon scenario, especially when the opponent is already statically sophisticated. Therefore, a warm start at a skillful initialization is essential for good efficiency. Meanwhile, the style of the initialized model is responsible for the style-of-play at convergence of training as we have seen in Section 6.2. Hence, the diversity in the pretrained model is responsible for the diversity of the population. For instance, Lei et al. [36] diversify their pretrained models to ensure the diversity of the exploiter policies. Inspired by this, we use the diverse population obtained from stage one as an initialization pool for the following experiments.

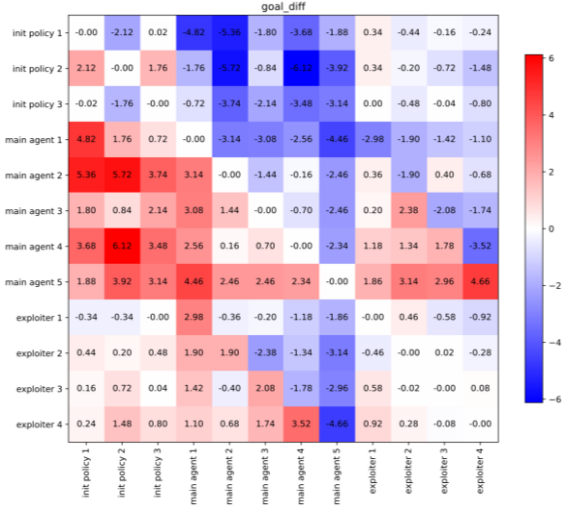

**(a)** The payoff table of a league training trial.

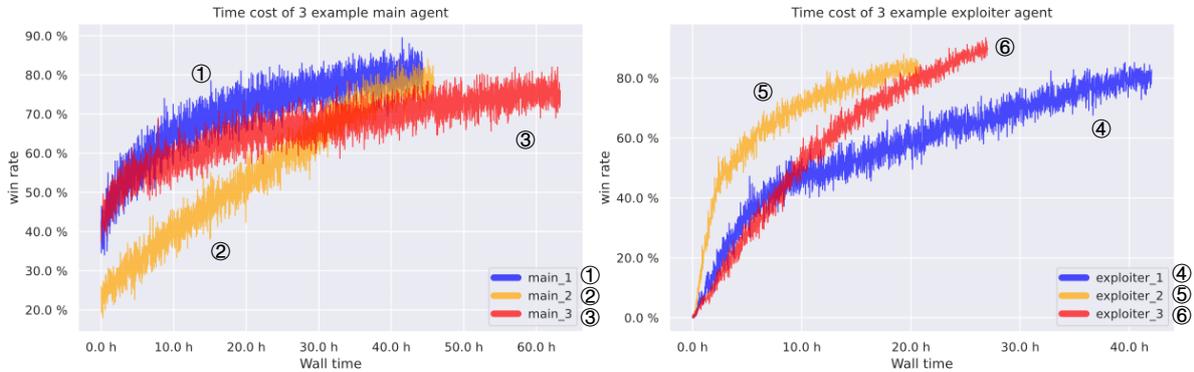

**(b)** The time cost of 3 example main and exploiter agents at later stage of league training. (①②③: main_agent_1,2&3; ④⑤⑥: exploiter_agent_1,2&3)

**Fig.16 (a)** The payoff table of a league training trial. The *init_policy 1,2 and 3* are the initial policy populations obtained from stage 2. Each entry represents the goal difference of the policy on the vertical axis against policy on the horizontal axis. The exploiter targets solely the newest main agent and the main agent needs to beat everyone in the population. The exploiter is also initialized from policies at stage 2. We see that the exploiter keeps exploiting the newest main agent and the main agent keeps learning to beat the new exploiter. **(b)** The time cost plot of 3 example main and exploiter agents at later stages of league training. Each generation of main and exploiter agents takes 1-2 days to reach an 80% win-rate.

**Opponents determine the learned strategy style.** Apart from the initialization diversity, we are also concerning about the style evolved during training. After an initialized policy has been set up, one way to guide policy learning is by reward shaping, as discussed in Section 6.1. However, according to our experience, we found that setting up a good reward shaping scheme for sophisticated styles of play is extremely difficult. For example, adding rewards for ball passing can increase the number of passes but most of them are unnecessary and can easily cause turnover. Adding penalties for turnovers will discourage passing and the trade-off is computationally hard to decide.

An easier and straightforward way is to choose the style of the opponent policy for the specific style-of-play we want the trained policy to possess. For example, when competing against a policy with a high-pressured defense, the policy can easily learn to pass the ball even under simple goal and lose goal reward settings. At this stage, we select a good initialization policy and further improve skills by choosing opponents with care. Our released pretrained policies *group_pressure* and *offensive_passer* are obtained in this manner. *Group_pressure* is trained using defensive reward shaping and *offensive_passer* learns good passing skills by training against *group_pressure* and some of its checkpoint policies.

To date, we have a population of policies with strategic diversity that are relatively less exploitable compared to stage one. In the final stage we deploy League Training to improve the general abilities of the population.

## 6.4 Stage Three: League Training

League training is similar to PSRO but has a more complex framework that has been used to beat professional human players in StarCraft [7]. L-MALib also provides league training framework implementation. In league training, a main-agent is trained constantly to keep improving its ability (similar to the main agent in PSRO with inherits) and the opponents are sampled following Prioritize Fictious Self-Play [7] (PFSP) which selects the opponents it cannot beat based on the win rate. Thus, the goal of the main agent is to beat everyone in the population rather than a game theory subset of policies, enlarging the scope of opponent style-of-play and improving general abilities. In the meantime, an exploiter policy is reset every generation and trained against the newest main agent only. Its job is to exploit the main-agent's strategic weakness and help the main-agent fix it by adding a copy of itself to the population. The exploiter can also potentially discover a new style-of-play as it does not inherit or only inherits from a general pretrained policy and has large space for evolving depending on the opponent. An example of a league training trial is shown in **Fig.16**. In **Fig.16 (b)**, we also provide plots of time cost at later stage of league training for reference purposes.

League training can incorporate ideas from both stages one and two. The exploiter can learn offensively and defensively using offense and defense rewards respectively. Each of them exploits the main agent from different strategic aspects. The offensive exploiter examines the main agent's defensive ability and vice versa. The exploiters can also adopt different initialization policies. The opponent selection mechanism of league training also ensures opponent diversity and encourages the main agent to grasp general skillsto beat everyone in the population.

We follow the three-stage procedures and obtain general football policies that are strong both offensively and defensively. We release some of the policies with representational style-of-play, serving as good initialization and a strong opponent to combat.

## 6.5 The Trained Policies

We release five representative trained-policies with diverse football tactics. Taking inspiration from statistical football tactics research [37,38], we evaluate their performance from 11 football-specific metrics using position data collected from policy simulations. Those metrics are: *Goal, Shot, Shot Accuracy, Ball Possession, Pass, Pass Accuracy, Tackles, Interceptions, Assist, Player Movement and Team Formation.* Most statistics are collected from event detection and the team formation is reviewed by *Weighted Team Centroid, Effective Player Space, Team Separateness and Player Length per Width.* A radar plot is in **Fig.17**.

Weighted Team Centroid is computed similarly to team centroid but weighted by distance between players and the ball; Effective Player Space (EPS) is computed as the area of the largest convex hull cover by the players; Team Separateness is the sum of distance between each player to its closest opponent. The Player Length per Width is the proportion of pitch length and the pitch width used by the player [37, 38].

Among all released policies, *beat_bot* is trained solely against the hard bot and exploits glitches in bots. *group_pressure* is obtained from defensive reward shaping at stage 1, which exerts high pressure on ball possession. The *offensive_passer* is obtained from stage 2 by carefully selecting a high pressure policy to encourage ball passing. The others are from league training where *defensive_passer* is one of the exploiter policies and *current_best* is the newest main agent. They learn to pass the ball and have better general skills, such as zone defense and cross-from-side tactics. This has shown the effectiveness of our 3-stage procedure in terms of building general football AI. We hope that these benchmark policies can serve as good pretrained models and good opponents to compete with for research purposes.

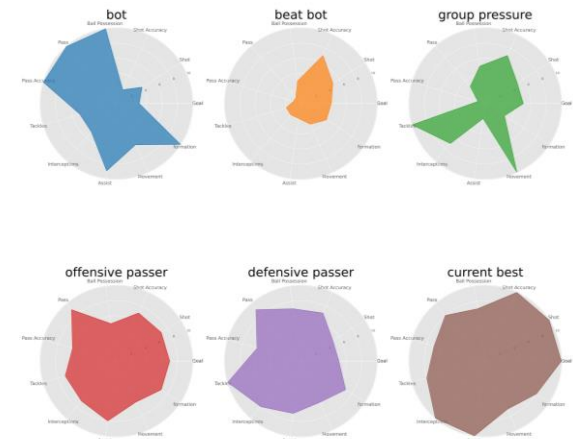

**Fig.17** Skill radar plots of the released policies and the 1.0 hard bot. The statistics are collected from matchup against each other and rescaled to be shown. *Group_pressure* is obtained from stage 1; *offensive_passer* is obtained from stage 2; *defensive_passer* and *current_best* are from stage 3 league training. We can see that the 3-stage procedure improves general skills of the policies.

## 7 Conclusions

In summary, we present a population-based multi-agent distributed RL framework and target specifically on Google Research Football testbeds. We provide an empirical study on 11v11 multi-agent full-game scenarios from scratch that are rarely explored and lack open-sourced evaluation. Our experiments show that the parameter-sharing independent PPO is sufficient to beat the bot with 1.0 difficulty but the resulting behavior is highly exploitable. We also present the population-based training pipeline to build strong football AI. All of our code and trained policies are released.

## Acknowledgements

This work was supported by the National Natural

# Appendix

## A.1 Notation

| | |
|---|---|
| State space | $\mathcal{S}$ |
| Action space | $\mathcal{A}$ |
| Observation space | $\mathcal{O}$ |
| Reward function | $R(\cdot)$ |
| Transition function | $P(\cdot)$ |
| Number of agents | $n$ |
| Discount value | $\gamma$ |
| Time step | $t$ |
| State at $t$ | $s_t$ |
| Joint action at $t$ | $A_t$ |
| Individual action at $t$ | $a_t$ |
| Observation at $t$ | $o_t$ |
| Policy function | $\pi(\cdot)$ |
| Policy function parameter | $\theta$ |
| Previous policy function parameter | $\theta_{old}$ |
| Value function | $V(\cdot)$ |
| Value function parameter | $\phi$ |
| Objective function | $\mathcal{J}(\cdot)$ |
| Expectation function | $\mathbb{E}(\cdot)$ |
| Loss function | $\mathcal{L}(\cdot)$ |
| State occupancy distribution | $\rho$ |
| Advantage function | $\hat{A}(\cdot)$ |
| Noise parameter | $\epsilon$ |
| Clipping operation | $clip$ |
| GAE lambda | $\lambda$ |
| TD error | $\delta$ |
| Entropy function | $\mathcal{H}(\cdot)$ |
| GAE return | $\hat{V}$ |
| Entropy coefficient | $\lambda_{entropy}$ |
| Set of policy profiles | $\Pi$ |
| Utility function | $U(\cdot,\cdot)$ |
| Compute best response policy | $BR$ |
| Payoff matrix | $\mathcal{M}$ |

**Table 3** Notation used in this paper

## A.2 L-MALib and MALib

MALib [2] integrates multiple benchmark environments and classical RL algorithms implementation, which we have omitted for simplicity. This means that MALib is a far more general framework while ours merely contains modules necessary for our experimental purpose. Our amendments can be summarized as follows:

1. L-MALib inherits the idea of centralized data server and policy server and enables much simpler data storage and adds limiters for various sampling schemes.

2. L-MALib optimizes and evaluates performance of asynchronous rollout and training (similar to IMPALA [32]) on multi-node and multi-GPU machines.

3. L-MALib adds many modifications to the GRF environment, such as statistics analysis, reward shaping, etc.

4. L-MALib makes policy architecture extensible for various settings, such as multi-headed, dimension extension, etc.

5. L-MALib adds a league training [7] framework.

## A.3 Environment Setting

We focus on 11v11 multi-agent scenarios with 10 players each team controllable[4] (excluding the goalkeepers) as illustrated in **Fig.18**. We mainly test on official *11_vs_11_hard_stochastic* and *11_vs_11_kaggle* scenarios. The observation is fully-observed for each agent, containing information on both teammates and opponents. The duration is 3000 corresponding to 90 minutes in the real world. Each teams on both sides obtains more scores than the other side and will win this game. Otherwise, both teams make a tie when the scores are even. In addition, we insert done flags at the moment when one team scores to segment the long-horizon trajectory.

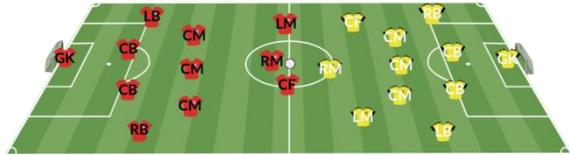

**Fig.18** Initial lineup of the 11_vs_11 scenarios where each team owns 11 players with 10 players controllable (except for the goalkeeper). In the figure the left team is at the kickoff.

## A.4 Experimental Setup

In the following subsections, we list the default configurations in our experiments.

---

[4] This setting aligns with Tikick [27]. The goalkeeper controlled by built-in logics can save the ball but when controlled by the agent its action space is aligned with other players and therefore cannot perform ball-saving anymore.

**Feature encoding**: The raw observation provided by Google Football is fully observable, containing information on player status from both teams, ball status, match state, etc. The official vectorized observation *simple115_v2* with dimension 115 provided by GRF is hard to learn (experiments shown in Section 5.1) as it only contains the absolute position and direction of players and the ball. In real life, a football player might also be interested in its close teammates and opponents. Thus, we additionally add features of relative position in this regard. In particular, we borrowed the basic version of features from *football-paris* [39] which enlarged the dimension up to 217. For each single player, this will include the relative position of the current player and the status of the closest teammates and the opponents as shown in **Table 4**. We name it *encoder_basic*. We also construct an enhanced version named *encoder_enhanced* which includes extra information regarding off-side and match states.

**Action masking**: Action masking is essential for speeding up convergence. We use action masking similar to *football-paris* [39] which throws away possible undesired moves depending on the current state. It can be summarized as follows:

1. When one of our players owns the ball, we disable other players' pass, shoot and dribble when the ball is far from them;
2. When the opponents own the ball, we disable a player's slide when the ball is far from it;
3. When nobody owns the ball, we disable a player's pass, shoot, dribble, and slide when the ball is far from it.
4. We disable a player's shoot when the ball is away from the opponent's penalty area and disable its long and high pass when within the penalty area.
5. Game-mode-specific action masking.

We use this action masking as it can accelerate training agents to beat built-in AI with hard difficulty 1.0. However, when facing more skilled opponents, agents without action masking are more likely to learn finer control and strategies. Therefore, we remove masking in population-based training.

**Reward shaping**: The football game is a sparse reward problem, where agents can only receive scoring rewards when goals. Based on prior experience with football, we designed several reward functions and traced each reward factor back to the time step when the corresponding event occured. For instance, the player receives an assisting reward the moment he starts passing the ball to his teammate who scores afterward. At default, we use simple zero-sum goal/lose-goal team rewards for experiments against built-in AI as we found it sufficient, and complex rewards settings for population-based training including individual rewards for assisting and losing/gaining balls to improve the skills of the policies.

Simple reward shaping is summarized in **Table 5**.

| Reward Factor | Description |
|---|---|

| Team goal/lose goal | Goal (+1) or lose a goal (-1) for all teammates scaled by their roles |
|---|---|
| Individual goal and assist | Player who goal (+1) and player who assist (+1) |
| Individual lose ball | Player who lose ownership of the ball (-1) |
| Individual gain ball | Player who gain ownership of the ball(+1) |

**Table 5** Simple reward shaping implemented.

| Feature Encoder | Encoded infos | Dimension |
|---|---|---|
| *encoder_basic* | **player_state:** left team position, direction, speed, roles, others | 217 |
| | **ball state:** ball position, zone, relative position, direction, speed, ownership | |
| | **left team state:** teammate relative position, direction, speed, distance | |
| | **right team state:** opponent relative position, direction, speed, distance | |
| | **left closest state:** state of the closest teammate | |
| | **right closest state:** state of the closet opponent | |
| | **available action:** action mask | |
| *encoder_enhanced* | **encoder_basic + offside infos + match state** | 312 |
| *simple115v2* | position and direction of left and right team+ball_state+ownership encoding+active player encoding+game mode encoding+available action | 115+1 9 |

**Table 4** Feature encoder implemented.

The complex reward shaping includes rewards for active offense, active defense and others. The active offense is designed to encourage attacks by:

1. *Goal difference*: the difference in goals between the left and right teams;
2. *Win reward*: the difference in goals but only given when winning the game;
3. *Ball position*: obtain a positive reward when the ball is at the opponent's half and the value increases if the ball is within the opponent's penalty area, which encourages the player to move the ball forwards;
4. *Goal pass*: the number of passes before a goal, reset if the opponent scores. This encourage passing that helps

to score;

5. *Shot reward*: rewarding shot when it leads to a goal;

To encourage active defense, we have designed the following:

1. *Minimum distance*: The negative distance between the player and the closed opponent, which results in high-pressure defense;
2. *Lost possession*: penalize if losing ball control;
3. *Get possession*: reward if steel or intercept the ball;
4. *Hold ball*: reward if we hold the ball and penalty if the opponent holds the ball;

Other reward shaping includes the following:

1. *Role-based*: reward goal and penalize goal lose according to player's role. The forwards obtains more goal rewards and less goal loss penalty, and the guard is the opposite. This helps players to play their corresponding roles.
2. *Passing*: reward good passing and penalize bad passing;

**Hyperparameter settings**: The hyperparameter setting contains configurations for rollout, training, data collecting and algorithm setup as shown in **Table 6**.

| | | |
|---|---|---|
| **Rollout** | rollout length | 3000 |
| | sample length | 1000 |
| | env | 11_vs_11 hard |
| | action set | default |
| **Training** | local queue size | 1 |
| | update interval | 1 |
| | optimizer | Adam |
| | actor_lr | 5e-4 |
| | critic_lr | 5e-4 |
| | optimizer eps | 1e-5 |
| **Data server** | capacity | 1000 |
| | sampler type | lumrf |
| **Algorithm** | initialization | orthogonal |
| | gain | 1 |
| | actor net | [256,128,64], ReLU |
| | critic net | [256,128,64], ReLU |
| | PPO epoch | 5 |
| | mini batch | 1 |
| | GAE lambda | 0.95 |
| | feature norm | True |
| | PopArt beta | 0.99999 |
| | entropy coefficient | 0.0001 |
| | clip value | 0.2 |
| | KL early-stopping | 0.01 |

**Table 6** Hyperparameter setting

## A.5 Experiments Across Seeds

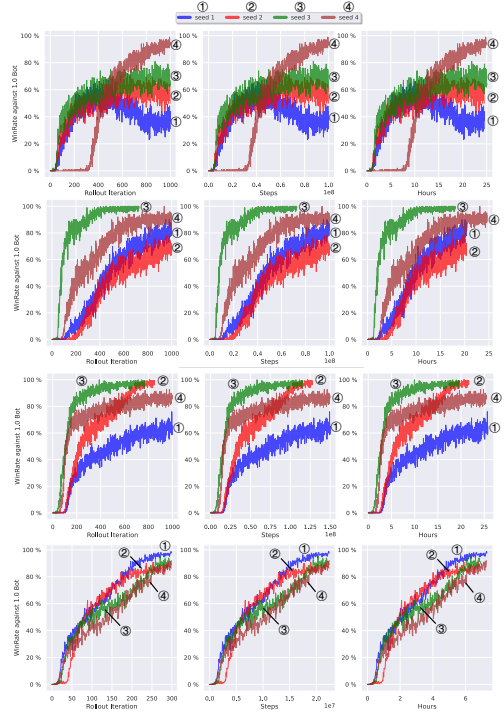

(a)    Discount factor $\gamma = 1$

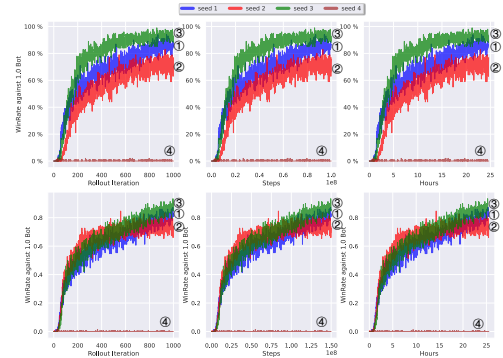

(b)    Discount factor $\gamma = 0.995$

**Fig.19** Performance in different seeds (1,2,3&4) at discount factors of 1 and 0.995. **(a)** Configurations from top to bottom denoted in the form [*rollout-worker, batch-size, discount-value*] are: [100,100,1]; [100,200,1]; [150,300,1]; [250,500,1] **(b)** The upper is [100,200,0.995] and the lower is [150,300,0.995]. (①②③④: seed 1,2,3&4.)

## A.6 State-Value Estimate $V$ and TD Error $\delta$ on Checkpointed Models

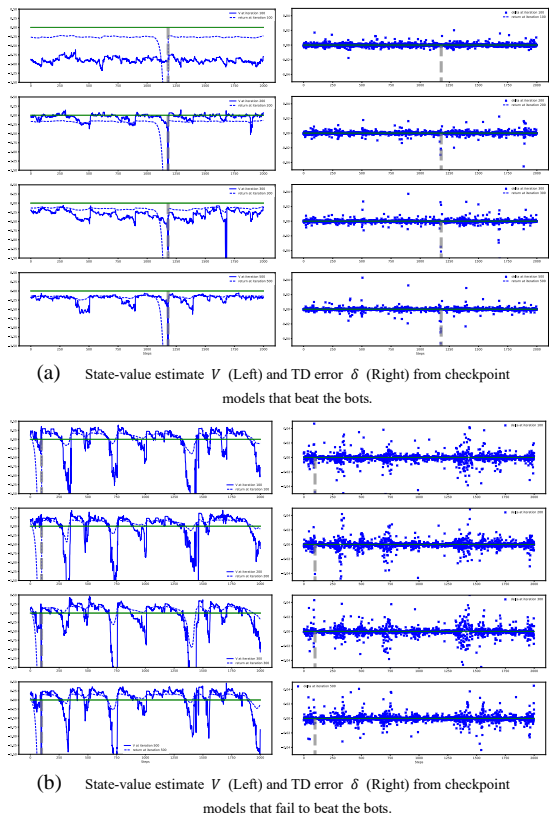

(a) State-value estimate $V$ (Left) and TD error $\delta$ (Right) from checkpoint models that beat the bots.

(b) State-value estimate $V$ (Left) and TD error $\delta$ (Right) from checkpoint models that fail to beat the bots.

**Fig.20** State-value estimate and TD error evaluation on a testing trajectory from (a) checkpoint models that beat the bot and (b) checkpoint models that fail to beat the bots. As the training proceeds, the models in (a) manage to learn a stable $V$ while the models in (b) experience large TD errors and learn a pessimistic $V$ which shows great aversion over unrewarding states.